\documentclass{article}
\usepackage{ijcai19}

\usepackage{url}
\usepackage{booktabs}
\usepackage{algorithm}
\usepackage{times}

\usepackage[utf8]{inputenc}
\usepackage[small]{caption}
\usepackage{multirow}
\usepackage{amsmath,amssymb,amsfonts}
\usepackage{graphicx}
\usepackage{epsfig}
\usepackage{diagbox}
\usepackage[skip=2pt]{caption}
\makeatletter
\newcommand*{\rom}[1]{\expandafter\@slowromancap\romannumeral #1@}
\makeatother

\newcolumntype{C}{>{\centering\arraybackslash} m{1.5cm} }
\makeatletter
\@namedef{ver@everyshi.sty}{}
\makeatother
\usepackage{algorithmic}
\usepackage{textcomp}
\usepackage{xcolor}

\newtheorem{definition}{Definition}
\DeclareMathOperator{\E}{\mathbb{E}}

\title{Generative Image Inpainting with Submanifold Alignment}

\author{
Ang Li\footnote{Corresponding authors.}\and
Jianzhong Qi\and
Rui Zhang\and
Xingjun Ma$^*$\And
Kotagiri Ramamohanarao\\
\affiliations
The University of Melbourne\\
\emails
angl4@student.unimelb.edu.au,
\{jianzhong.qi, rui.zhang, xingjun.ma, kotagiri\}@unimelb.edu.au
}

\begin{document}

\maketitle

\begin{abstract}
Image inpainting aims at restoring missing regions of corrupted images, which has many applications such as image restoration and object removal.
However, current GAN-based generative inpainting models do not explicitly exploit the structural or textural consistency between restored contents and their surrounding contexts.
% For example, given an image of a person wearing sunglasses where one side of the sunglasses is missing, current models may incorrectly restore the missing sunglass with a human eye.
To address this limitation, we propose to enforce the alignment (or closeness) between the local data submanifolds (or subspaces) around restored images and those around the original (uncorrupted) images during the learning process of GAN-based inpainting models.
We exploit Local Intrinsic Dimensionality (LID) to measure, in deep feature space, the alignment between data submanifolds learned by a GAN model and those of the original data, from a perspective of both images (denoted as iLID) and local patches (denoted as pLID) of images. We then apply iLID and pLID as regularizations for GAN-based inpainting models to encourage two levels of submanifold alignment: 1) an image-level alignment for improving structural consistency, and 2) a patch-level alignment for improving textural details.
% The dimensional properties of a data submanifold in the vicinity of a reference point can be measured by an expansion-based dimensionality measure called Local Intrinsic Dimensionality (LID) in the deep feature space of a GAN discriminator network. In this paper, we extend the LID measure to define the alignment between two data submanifolds, and develop image-level and patch-level submanifold alignment regularizations to GAN-based inpainting models.
% To address this limitation, our key insight is that understanding from the perspective of Local Intrinsic Dimensionality (LID) can provide essential guidance for the inpainting process.
% Given the distance distribution of a reference example to its neighbors, LID evaluates the space-filling ability of the example's surrounding region.
% Based on this insight, we propose a dimensionality-driven image inpainting model that utilizes the combination of two LID-based regularizations: an Image-Based Average LID to enhance global semantic consistency and a Patch-Based Average LID to ensure semantic consistency of local details.
Experimental results on four benchmark datasets show that our proposed model can generate more accurate results than state-of-the-art models.
\end{abstract}

\section{Introduction}\label{secIntro}
\begin{figure}[t!]
\centering
\setlength{\tabcolsep}{0.2em}
{
\begin{tabular}{cccc}

\includegraphics[width=2cm]{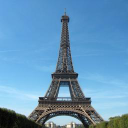}&
\includegraphics[width=2cm]{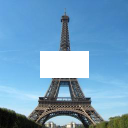}&
\includegraphics[width=2cm]{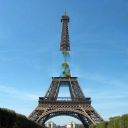}&
\includegraphics[width=2cm]{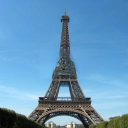}
\\

\includegraphics[width=2cm]{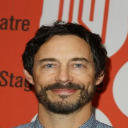}&
\includegraphics[width=2cm]{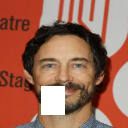}&
\includegraphics[width=2cm]{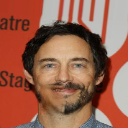}&
\includegraphics[width=2cm]{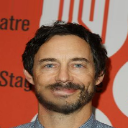}
\\

\small(a) Original &\small (b) Corrupted &\small (c) GMCNN  &\small (d) Ours\\

\end{tabular}
}
\caption{Comparison between the state-of-the-art GMCNN and our proposed model.}
\label{exampleCompare} % I can do without the label too
% \vspace{-0.5cm}
\vspace{-0.11 in}
\end{figure}

Given a corrupted image where part of the image is missing, image inpainting aims to synthesize plausible contents that are coherent with non-missing regions.
Figure \ref{exampleCompare} illustrates such an example, where Figure 1a and 1b show two original (uncorrupted) images and their corrupted versions with missing regions respectively.
The aim is to restore the missing regions of corrupted images in Figure 1b, such that the restored regions contain contents that make the whole image look natural and undamaged (high structural and textural consistency with the original images in Figure 1a). 
With the help of image inpainting, applications such as restoring damaged images or removing blocking contents from images can be realized.

Generative Adversarial Networks (GANs) are powerful models for image generation \cite{goodfellow2014generative,radford2016unsupervised}.
From the perspective of GANs, image inpainting can be viewed as a conditional image generation task given the context of uncorrupted regions \cite{pathak2016context,iizuka2017globally,Yang_2017_CVPR,yeh2017semantic,yu2018generative,Wang2018Image}.
A representative GAN-based inpainting model is Context Encoder \cite{pathak2016context}, where a convolutional encoder-decoder network is trained with the combination of a reconstruction loss and an adversarial loss \cite{goodfellow2014generative}.
% to predict the missing contents.
The reconstruction loss guides to recover the overall coarse structure of the missing region, while the adversarial loss guides to choose a specific distribution from the results to promote natural-looking patterns.
Lately, Wang \emph{et al.} \shortcite{Wang2018Image} propose a generative multi-column neural network (GMCNN) for better restoration.

In spite of the encouraging results, the restored images by current GAN-based inpainting models may lead to \emph{structural or textural inconsistency} compared with the original images.
Specifically, current inpainting models may significantly alter the structure or the texture of the objects involved in the missing regions, and therefore the restored contents visually mismatch their surrounding undamaged contents.
%current GAN-based inpainting models do not explicitly consider the image content consistency in terms of semantics.
%Specifically, the inpainting process may ignore semantic information of the non-missing context such as attribute information and segmentation information.
%Therefore, the restored images may be perceptually contradictory with the natural images. 
For example, in Figure 1c, the restored regions by the state-of-the-art inpainting model GMCNN \cite{Wang2018Image} exhibit segmentation structure misalignment (part of the tower is missing) in the top image and texture mismatch (inconsistent with respect to `beard' pattern) in the bottom image.
This problem can be ascribed to the following two limitations.

First, GAN-based formulation has the ill-posed nature: the one-to-many mapping relationships between a missing region and its possible restorations.
This can lead to sub-optimal inpainting results as in the above examples, especially when the inpainting model was not properly regularized for structural/textural consistency.
% due to the lack of specific regularizations on semantic consistency during inpainting model training.
% The commonly used loss function with a reconstruction loss for pixel-wise identity and an adversarial loss for determining overall image realness, however, fails to address this limitation.
The commonly used loss combination of a reconstruction loss and an adversarial loss, however, fails to address this limitation.
The reconstruction loss mainly focuses on minimizing pixel-wise difference (very low-level), while the adversarial loss determines the overall similarity to real images (very high-level).
Intuitively, mid-level (between the pixel and the overall realness levels) regularizations that can provide more explicit emphases on both structural and textural consistencies are needed for natural restorations.

Second, most existing inpainting models \cite{Yang_2017_CVPR,yu2018generative} adopt a two-stage ``coarse-to-fine'' inpainting process.
Specifically, these models generate a coarse result at the first stage (which also suffers from the first limitation), then refine the coarse result at the second stage by replacing patches in the restored regions with their nearest neighbors found in the original patches. Due to the cosine-similarity based searching of nearest patches, these models tend to produce repetitive patterns.
% However, these models will lead to smooth and repetitive patterns, and therefore reduce structure diversity, due to the cosine similarity based searching of nearest natural patches.
Moreover, when structures in missing regions are originally different from those in the background, these models may significantly alter or distort the structure of the restored contents and end up with unrealistic restorations, for example in the top image of Figure 1c, filling missing part of the `tower' with `sky' background.

To address the two limitations, we propose to enforce the alignment (or closeness) between the local data submanifolds (or subspaces) around restored images and those around the original images, during the learning process of GAN-based inpainting models. 
Our intuition is that, a restored image or patch of the image will look more natural if the image or the patch is drawn from a data submanifold that is closely surrounding the original image or patch, that is, data submanifolds around the restored images/patches align well with those around the original images/patches. 
In this paper, we adapt an expansion-based dimensionality measure called Local Intrinsic Dimensionality (LID) to characterize the dimensional property of the local data submanifold (in deep feature space) around a reference image/patch. 
We further exploit LID to measure the \emph{submanifold alignment} from a perspective of either images or local patches. This allows us to develop two different levels of submanifold alignment regularizations to improve GAN-based inpainting models: 1) an image-level submanifold alignment (iLID) for structural consistency, and 2) a patch-level submanifold alignment (pLID) for improving textural details. 
In summary, our main contributions are:
\begin{itemize}
  \item We propose and generalize the use of Local Intrinsic Dimensionality (LID) in image inpainting to measure how closely submanifolds around restored images/patches align with those around the original images/patches.
  
  \item With the generalized LID measure, we develop two submanifold alignment regularizations for GAN-based inpainting models to improve structural consistency and textural details.
  
  \item Experiments show that our model can effectively reduce structural/textural inconsistency and achieve more accurate inpainting compared with state-of-the-art models.
\end{itemize}

\section{Related Work}\label{secBackground}

% \subsection{Generative Adversarial Networks}
% Generative Adversarial Network (GAN) \cite{goodfellow2014generative} consists of two competitively learning networks: a generator network and a discriminator network. The discriminator examines data samples to identify whether they are from the generator's distribution or the training data distribution, while the generator learns to fool the discriminator by generating samples that share similar distribution with the training data. The two networks are jointly trained to optimize an objective function resembling a two-player minimax game.
% % Both networks are trained alternatively and the competition drives them to improve until the generated samples are indistinguishable from the genuine samples.
% The original GAN model is an unconditioned generative model which lacks control on modes of the data being generated.
% Later, Mirza and Osindero \cite{Mirza2014Conditional} propose a conditional GAN based on additional information such as class labels or data from other modalities to direct the data generation process.
% Radford \emph{et al.} \cite{radford2016unsupervised} propose DCGAN to combine Convolutional Neural Network (CNN) with GAN. The DCGAN architecture has been widely used by GAN-based image inpainting models.

% \subsection{Image Inpainting}
Traditional image inpainting models are mostly based on patch matching \cite{barnes2009patchmatch,bertalmio2003simultaneous} or texture synthesis \cite{efros1999texture,efros2001image}.
They often suffer from low generation quality, especially when dealing with large arbitrary missing regions.
Recently, deep learning and GAN-based approaches have been employed to produce more promising inpainting results \cite{pathak2016context,Yang_2017_CVPR,yeh2017semantic,iizuka2017globally,yu2018generative,liu2018image,song2018contextual,zhang2018semantic,Wang2018Image,Ang_2019_IJCNN}.
Pathak \emph{et al.}\shortcite{pathak2016context} first propose the \textit{Context Encoder} (CE) model that has an encoder-decoder CNN structure.
They train CE with the combination of a reconstruction loss and an adversarial loss \cite{goodfellow2014generative}.
%The CE model can generate better images comparing with those generated by traditional models.
%However, the images generated by CE still tend to be blurry with evident artifacts and lack fine-grained details due to the limitation of the basic encoder-decoder generator structures.
Later models apply post-processing on images produced by encoder-decoder models to further improve their quality.
Yang \emph{et al.} \shortcite{Yang_2017_CVPR} propose one such model that adopts CE at its first stage, while at a second stage, they refine the inpainting results by propagating surrounding texture information.
Iizuka \emph{et al.} \shortcite{iizuka2017globally} propose to use both global and local discriminators at the first stage, and Poisson blending at the second stage.
Yu \emph{et al.} \shortcite{yu2018generative} propose a refinement network for post-processing based on contextual attention.
These models mainly focus on enhancing the resolution of inpainted images, while ignoring the structural or textural consistency between the restored contents and their surrounding contexts.
These models are all based on an encoder-decoder model for generating initial results, which can easily suffer from the ill-posed one-to-many ambiguity. This may produce suboptimal intermediate results that limit the effectiveness of post-processing at a later stage. 
Lately, Wang \emph{et al.} \shortcite{Wang2018Image} propose a generative multi-column neural network (GMCNN) for better restoration of global structures, however, as we show in Figure \ref{exampleCompare}c, it still suffers from the structural or textural inconsistency problem.

% \subsection{Dimensionality and Manifold Learning}
% The Local Intrinsic Dimensionality (LID) model \cite{Houle2017a} was recently used for successful detection of adversarial examples for DNNs by \cite{Ma2018a}. This work demonstrates that adversarial perturbations (one type of input noise) tend to increase the dimensionality of the local subspace immediately surrounding a test sample, and that features based on LID can be used for identifying such perturbations.
% However, in this paper we show how LID can be used in a new way, as a tool for assessing the learning behavior of a DNN, and developing an adaptive learning strategy against noisy labels.

% Other works have also considered the use of dimensionality measures for regularization in manifold learning \cite{roweis2000nonlinear,belkin2004regularization,belkin2006manifold}. For example, an intrinsic geometry regularization over Reproducing Kernel Hilbert Spaces (RKHS) was proposed in \cite{belkin2006manifold} to enforce smoothness of solutions relative to the underlying manifold, and a Laplacian-based regularization using the weighted neighborhood graph was proposed in \cite{belkin2004regularization}. In contrast to these works, which treated dimensionality as a characteristic of the global data distribution, we explore how knowledge of local dimensional characteristics can be used to monitor and modify DNN learning behavior for the noisy label scenario. 
% !TEX encoding = UTF-8 Unicode
\section{Dimensional Characterization of Local Data Submanifolds}\label{secLid}
We start with a brief introduction of the Local Intrinsic Dimensionality (LID) \cite{Houle2017a} measure for assessing the dimensional properties of local data submanifolds/subspaces.
% We then discuss some recent applications that apply LID to characterize data subspaces residing in the deep feature space of deep neural networks (DNNs).

\subsection{Local Intrinsic Dimensionality}
LID measures the rate of growth (\textit{e.g.} expansion rate) in the number of data points encountered as the distance from a reference point increases. Intuitively, in Euclidean space, the volume of an $D$-dimensional ball grows proportionally to $r^D$ when its size is scaled by a factor of $r$. From the above rate of volume growth with distance, the dimension $D$ can be deduced from two volume measurements as:
\begin{equation}
V_2/V_1 = (r_2/r_1)^D \Rightarrow D = \ln(V_2/V_1)/\ln(r_2/r_1).   
\end{equation}
Transferring the concept of expansion dimension from the Euclidean space to the statistical setting of local distance distributions, the notion of $r$ becomes the distance from a reference sample $x$ to its neighbours and the ball volume is replaced by the cumulative distribution function of the distance. This leads to the formal definition of LID \cite{Houle2017a}.

% The LID can be viewed as the degree of a polynomial function with the most suitable fit to it, when retrieved among an infinitesimally small neighborhood domain of the origin \cite{Houle2017a}.
% Evaluation of \textit{Local Intrinsic Dimensionality} (LID) characteristics has benefited supervised learning recently, such as estimating the complexity of search queries in approximate similarity search \cite{casanova2017}, detecting adversarial samples for deep neural networks \cite{Ma2018a}, measuring the outlierness of data \cite{Houle2018}, and preventing overfitting for noisy label classification by deep neural networks.

\begin{definition}[Local Intrinsic Dimensionality] \quad \\
Given a data sample $x \in X$, let $r>0$ be a random variable denoting the distance from $x$ to other data samples. If the cumulative distribution function $F(r)$ is positive and continuously differentiable at distance $r>0$, the LID of $x$ at distance $r$ is given by:
\begin{equation} \label{eq:LID_r}
  \begin{split}
    \textup{LID}_F(r) & \triangleq \lim_{\epsilon\to 0} \frac{\ln\big(F((1+\epsilon) r)\big/F(r)\big)}{\ln((1+\epsilon)r/r)} 
     = \frac{r F'(r)}{F(r)},
  \end{split}
\end{equation}
whenever the limit exists.
\label{def:lid}
The \textup{LID} at $x$ is in turn defined as the limit when the radius $r \to 0$: 
\begin{equation} \label{eq:LID}
    \textup{LID}_F = \lim_{r \to 0}  \textup{LID}_F(r).
\end{equation}
\end{definition}

The last equality of Equation~\eqref{eq:LID_r} follows by applying L'H\^{o}pital's rule to the limits. $\text{LID}_F$ describes the relative rate at which the cumulative distance function $F(r)$ increases as the distance $r$ increases. In the ideal case where the data in the vicinity of $x$ are distributed uniformly within a local submanifold, $\text{LID}_F$ equals the dimension of the submanifold. In more general cases, LID provides a rough indication of the dimension of the submanifold containing $x$ that would best fit the data distribution in the vicinity of $x$.

\subsection{Estimation of LID}
There already exists several estimators of LID \cite{amsaleg2015estimating,Amsaleg2018Extreme,levina2005maximum}, among which the Maximum Likelihood Estimator (MLE) shows the best trade-off between statistical efficiency and complexity:
\begin{equation}
\small
\text{LID}\,(x; X) =
-\Big (\frac{1}{k} \sum_{i=1}^{k} \ln\frac{r_i(x; X)}{r_{max}(x; X)} \Big )^{-1}. \\
\label{eq:lidmle}
\end{equation}
where $k$ is the neighborhood size, $r_i(x; X)$ is the distance from $x$ to its $i$-th nearest neighbor in $X\setminus\{x\}$, and $r_{\textrm{max}}(x; X)$ denotes the maximum distance within the neighborhood (which by convention can be $r_k(x;X)$).

LID has recently been used in adversarial detection \cite{Ma2018a} to characterize adversarial subspaces of deep networks. 
It has also been applied to investigate the dimensionality of data subspaces  in the presence of noisy labels \cite{ma2018d}. 
In a recent work \cite{barua2019quality}, LID was exploited to measure the quality of GANs in terms of the degree to which manifolds of real data distribution and generated data distribution coincide with each other.
% Other applications of LID have also been seen in similarity search~(\cite{lidsearchquery}) and outlier detection \cite{Houle2018}. \cite{amsaleg2017vulnerability} \cite{houle2015r}
Inspired by these studies, in this paper, we develop the use of LID in image inpainting to encourage submanifold alignment for GAN-based generative inpainting models.

% !TEX encoding = UTF-8 Unicode
\section{Proposed Inpainting Model with Submanifold Alignment}\label{secApproach}

\begin{figure}[!bt]
%\vspace{-0.1 in}
\centering
\includegraphics[width=0.8\linewidth]{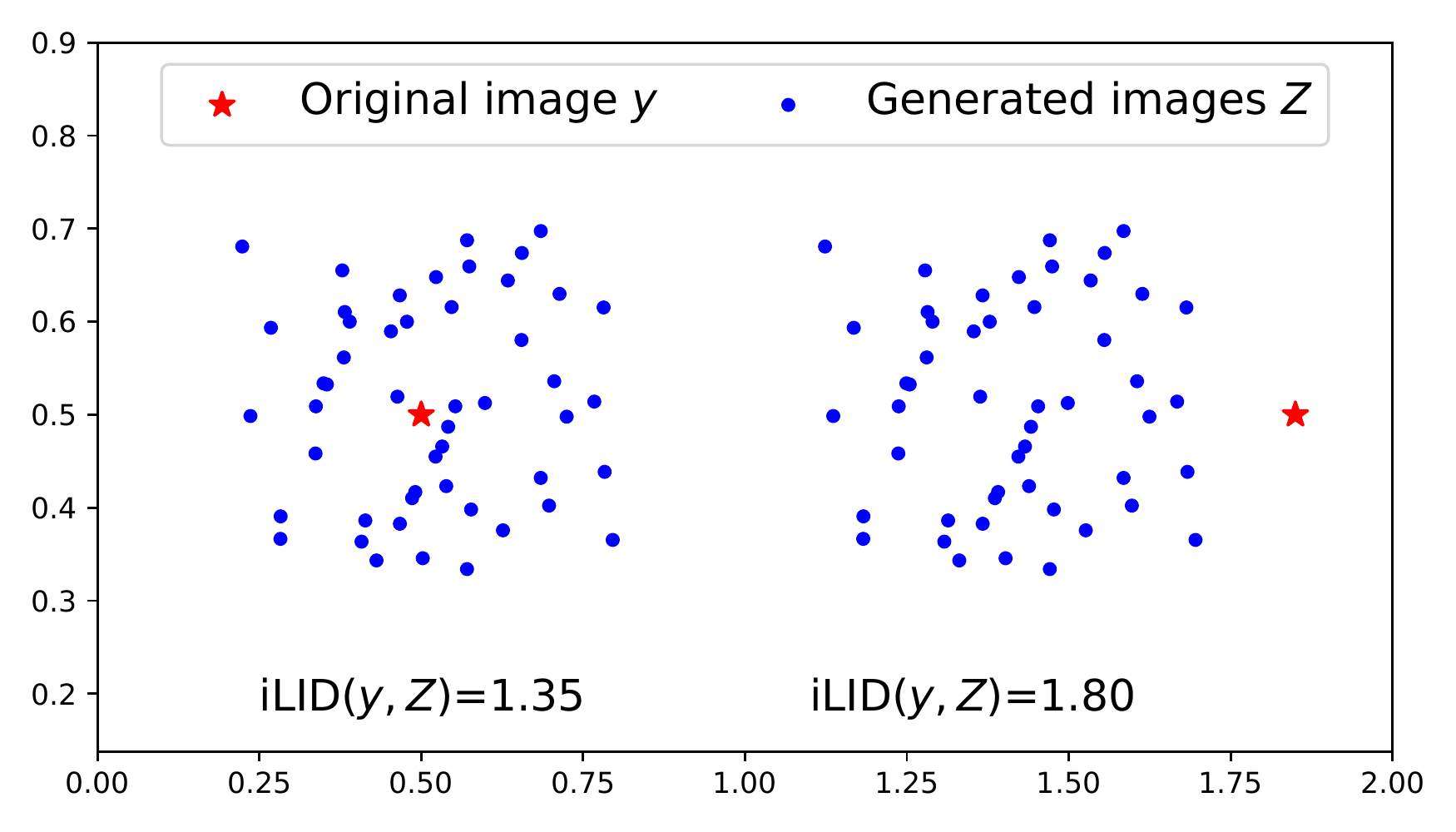}
\caption{This example shows how $\text{iLID}(y, Z)$ can reflect the closeness of the data submanifold of restored images $Z$ (blue points) with respect to an original image $y$ (red star).}
\label{fig:lid_toy}
%\vspace{-0.15 in}
\end{figure}

Given an input (corrupted) image $x \in X$ with missing regions, an image inpainting model aims to restore the missing regions so that the output (restored) image $z \in Z$ matches the ground truth (original) image $y \in Y$. Following previous works, we adopt GANs to generate the output image $z$. To address the structural and textural inconsistencies exist in current GAN-based inpainting models, we propose to enforce the restored images or local patches to be drawn from data submanifolds that align closely with submanifolds of the original images or patches. To achieve this, we develop the LID characterization of local data submanifolds to derive two levels of submanifold alignments: image-level and patch-level.
% In this work, we aim to address the aforementioned semantic inconsistency issue of image inpainting.
% Our insight is that if the generated samples' distribution and the real samples' distribution are similar, then the average spatial distance between every generated sample and its corresponding neighborhood of real samples should be low enough.
% Therefore, our scheme is to apply LID-based regularization in the training process, named \textit{Dimensionality-Driven Image Inpainting} (DimIpt).

\subsection{Image-level Submanifold Alignment}\label{sec:iLID}
We propose \textit{Image-level Local Intrinsic Dimensionality} (iLID) to measure how close data submanifolds around restored images are to those around the original images. For an original image, we define the iLID of its local submanifold by its neighbors found in GAN generated samples:
\begin{equation}\label{eq:lid_yz}
\text{iLID}(y; Z) = - \left (\frac{1}{k_{I}}\sum_{i=1}^{k_{I}}\log\frac{r_{i}(\phi(y),\phi(Z))}{r_{max}(\phi(y),\phi(Z))} \right )^{-1},
\end{equation}
where, transformation $\phi$ represents the $L_{th}$ layer of a pretrained network mapping raw images into deep feature space, $r_{i}(\phi(y),\phi(Z))$ is the $L_2$ distance of $\phi(y)$ to its $i$-th nearest neighbor $\phi(z_{i})$ in $\phi(Z)$, and $k_{I}$ is the neighborhood size.

Figure \ref{fig:lid_toy} illustrates a toy example showing how $\text{iLID}(y; Z)$ works. In the left scenario, GAN learns a data submanifold (with blue points indicating samples $Z$ drawn from this submanifold) that is closely surrounding the original image $y$ and the $\text{iLID}(y; Z)$ score is low, while in the right scenario, as the GAN-learned data submanifold drifts away from $y$ the $\text{iLID}(y; Z)$ score increases from 1.53 to 3.78. Note that, from the original LID perspective, the right scenario can be interpreted as a sample $y$ lies in a higher dimensional space that is off the normal data submanifold. Revisit Eq. \eqref{eq:lid_yz}, assuming $r_{max} > r_i$, as the distance between $y$ and its nearest neighbors in $Z$ increases by $d > 0$, then the term inside of the log function becomes $(r_i + d)/(r_{max} +d)$. Since,
\begin{equation}
    \frac{r_i + d}{r_{max} + d} - \frac{r_i}{r_{max}} = \frac{r_{max} - r_i}{r_{max}(r_{max}/d + 1)},
\end{equation}
iLID$(y; Z)$ will increase with the increase of $d$, reflecting the alignment/closeness drift of local data submanifold from $y$.

Taking the expectation of iLID$(y; Z)$ over all the original samples in $Y$ gives us the regularization for image-level submanifold alignment:
\begin{equation}
\mathcal{L}_{\text{iLID}}(Y; Z) = \E_{y \in Y}\text{iLID}(y; Z).
\end{equation}
Low $\mathcal{L}_{\text{iLID}}(Y; Z)$ scores indicate low average spatial distance between the elements of $Y$ and their neighbor sets in $Z$. $\mathcal{L}_{\text{iLID}}(Y; Z)$ provides a local view of how well data submanifolds learned by a GAN model align with the original data submanifolds. By optimizing $\mathcal{L}_{\text{iLID}}(Y; Z)$, we can encourage a GAN generator to draw restorations from data submanifolds that are closely surrounding the original images, thus can avoid structural inconsistency between the restored image and the original image.

%The feature extraction network is an ImageNet-pretrained VGG-19 network \cite{Simonyan2014Very}, and we take its output at the third convolutional layer (\textit{e.g.} $\phi$) as deep features. 
As computing neighborhoods for each $y \in Y$ with respect to large sets $Z$ can be prohibitively expensive, we use a batch-based estimation of LID: for one batch during training, we use the GAN generated samples as the $Z$ for that batch  to compute iLID score for each original sample in the batch. In other words, the $\mathcal{L}_{\text{iLID}}(Y; Z)$ is efficiently defined over a batch of samples (original or restored). 

\subsection{Patch-level Submanifold Alignment}
To avoid local textural inconsistency, we further propose the \textit{Patch-level Local Intrinsic Dimensionality} (pLID) based on feature patches in the deep feature space (rather than the raw pixel space). Feature patches were extracted on the feature map obtained by the same transformation $\phi$ as used for image-level submanifold alignment. Specifically, for an original image $y \in Y$ and its corresponding restored image $z \in Z$ with restored region $z^{o}$, we extract one set $P$ of $3 \times 3$ feature patches from the entire feature map $\phi(y)$, and one set $Q$ of $3 \times 3$ feature patches only from part of the feature map $\phi(z)$ that is associated with region $z^{o}$. Following similar formulation of iLID, we define pLID as:
\begin{equation}
\text{pLID}(p; Q) = - \left (\frac{1}{k_{P}}\sum_{i=1}^{k_{P}}\log\frac{r_{i}(p,Q)}{r_{max}(p,Q)} \right )^{-1},
\end{equation}
where, $p \in P$ is a patch in P and $k_{P}$ is the neighborhood size. Then, the patch-level submanifold alignment regularization can be defined as:
\begin{equation}
\mathcal{L}_{\text{pLID}}(Y; Z) = \E_{y \in Y}\E_{p \in P}\text{pLID}(p; Q).
\end{equation}

Optimizing $\mathcal{L}_{\text{pLID}}(Y; Z)$ enforces the GAN model to draw patches from data submanifolds that are closely surrounding the original patches, which helps improve texture consistency. Specifically, $\mathcal{L}_{\text{pLID}}(p; Q)$ encourages $p$ to receive reference from different neural patches in $Q$, thus can avoid repetitive patterns and maintain variety in local texture details. Similar to $\mathcal{L}_{\text{iLID}}(Y; Z)$ , we define the $\mathcal{L}_{\text{pLID}}(Y; Z)$ over a batch of original or restored samples during training. 

\subsection{The Overall Training Loss}
The overall training loss of our model is a combination of an adversarial loss, a reconstruction loss and the two proposed regularizations:
\begin{equation}
% \begin{split}
\mathcal{L} = \lambda_{I}\mathcal{L}_{\text{iLID}} + \lambda_{P}\mathcal{L}_{\text{pLID}} + \lambda_{A}\mathcal{L}_{\text{adv}} + \mathcal{L}_{\text{rec}},
% \end{split}
\end{equation}
where $\mathcal{L}_{\text{adv}}$ and $\mathcal{L}_{\text{rec}}$ denote the adversarial loss and reconstruction loss respectively, and parameters $\lambda_{I}$, $\lambda_{P}$ and $\lambda_{A}$ control the tradeoff between different components of the loss. Following recent works \cite{liu2018image,yu2018generative}, we adopt the improved Wasserstein GAN \cite{Gulrajani2017} as our adversarial loss:
\begin{equation}
\mathcal{L}_{\text{adv}} = -\E_{x \sim \mathbb{P}_{x}}[D(G(x))] + \lambda_{gp}\E_{\hat{x} \sim \mathbb{P}_{\hat{x}}}[(\| \bigtriangledown_{\hat{x}}D(\hat{x}) \| - 1)^{2}],
\end{equation}
where $G$ and $D$ denote the generator and the discriminator of our model respectively, $x \in X$ is an input (corrupted) sample and $z=G(x)$ is the GAN generated sample. The restored image for $x$ is $\hat{x} = tx + (1-t)G(x)$ with the mask $t \in \{0, 1\}$ having zeros at the missing pixels and ones at the non-missing pixels, and we use the $L_2$ reconstruction loss:
\begin{equation}
\mathcal{L}_{\text{rec}} = \E_{x \sim \mathbb{P}_{x}, y \sim \mathbb{P}_{y}}\| \hat{x} - y \|_{2}.
\end{equation}
%We denote our model as \emph{DimIpt} and our code is available for download at \url{https://www.overleaf.com/project/5c6cc227d1ba83158575249a}.

% We adopt the structure of Iizuka's work \cite{iizuka2017globally} as our model structure which is a GAN structure featured by dilated convolutions in generator and using a combination of global and local discriminators.
% With the proposed ImLID regularization and PaLID regularization, the objective function of our model can be formulated as:
% \begin{equation}
% \begin{split}
% \mathcal{L} = & \lambda_{Im}\mathcal{L}_{ImLID} + \lambda_{Pa}\mathcal{L}_{PaLID} \\
% & + \lambda_{adv}\mathcal{L}_{adv} + \mathcal{L}_{re}.
% \end{split}
% \end{equation}
% where $\mathcal{L}_{adv}$ and $\mathcal{L}_{re}$ denotes adversarial loss and reconstruction loss.
% Tradeoff parameters $\lambda_{Im}$, $\lambda_{Pa}$ and $\lambda_{adv}$ control the importance of corresponding terms.
% Following recent works \cite{yu2018generative}\cite{liu2018image}, we adopt the improved Wasserstein GAN \cite{Gulrajani2017} as our adversarial loss $\mathcal{L}_{adv}$:
% \begin{equation}
% \mathcal{L}_{adv} = -E_{x \sim \mathbb{P}_{x}}[D(G(x))] + \lambda_{gp}E_{\hat{x} \sim \mathbb{P}_{\hat{x}}}[(\| \bigtriangledown_{\hat{x}}D(\hat{x}) \| - 1)^{2}].
% \end{equation}
% where $G$ and $D$ denote generator and discriminator of our model, respectively.
% $x \in X$ is an input sample for our model.
% $\hat{x} = tG(x) + ty$ and $t \in [0, 1]$.
% Furthermore, we utilize $l_{2}$ distance as the reconstruction loss $\mathcal{L}_{re}$ to minimize pixel-wise difference.
% !TEX encoding = UTF-8 Unicode
\section{Experiments}\label{secExperiments}
We compare with four state-of-the-art inpainting models:  \textbf{CE} \cite{pathak2016context}, \textbf{GL} \cite{iizuka2017globally}, \textbf{GntIpt} \cite{yu2018generative} and \textbf{GMCNN} \cite{Wang2018Image}. 
We apply the proposed training strategy on both CE and GL GAN-architectures, and denote their improved versions as \textbf{CE+LID} and \textbf{GL+LID} respectively.
We evaluate all models on four benchmark datasets: Paris StreetView \cite{doersch2012makes}, CelebA \cite{liu2015deep}, Places2 \cite{zhou2017places} and ImageNet \cite{russakovsky2015imagenet}.
No pre-processing or post-processing is applied in all experiments.
% Both quantitative and qualitative evaluations are performed in this section.

\begin{table*}[hbt]
\renewcommand\arraystretch{1.05}
\begin{center}
\scalebox{0.95}
{
\begin{tabular}{lccccccccc}
      			\hline
      	 \multirow{2}{*}{Model}  & \multicolumn{2}{c}{Paris StreetView} & \multicolumn{2}{c}{CelebA} & \multicolumn{2}{c}{Places2} & \multicolumn{2}{c}{ImageNet} \\ 
     			& PSNR & SSIM  & PSNR & SSIM  & PSNR & SSIM  & PSNR & SSIM\\ \hline 
		CE \cite{pathak2016context} & 22.92 & 0.858 & 21.78  & 0.923  & 18.72 & 0.843 & 20.43 & 0.787 \\	 
      		GL \cite{iizuka2017globally} & 23.64 & 0.860 & 23.19  & 0.936  & 19.09 & 0.836 & 21.85 & 0.853 \\ 
		GntIpt \cite{yu2018generative} & 24.12 & 0.862 & 23.80 & 0.940 & 20.38 &  0.855 & 22.82 & \textbf{0.879} \\ 
      		GMCNN \cite{Wang2018Image} & 23.82 & 0.857 & 24.46 & 0.944 & 20.62 &  0.851 & 22.16 & 0.864 \\ \hline
      		\textbf{Ours (CE+LID)} & 23.78 & \textbf{0.873} & 24.05 & 0.947 & 20.17 &  0.848 & 22.11 & 0.860 \\ 
      		\textbf{Ours (GL+LID)} & \textbf{24.33} & 0.867 &  \textbf{25.56}  & \textbf{0.953} & \textbf{21.02} & \textbf{0.864} & \textbf{23.50} & 0.875 \\ \hline 
\end{tabular}    
}
\end{center}
%\vspace{-0.1 in}
\caption{Quantitative comparisons on Paris Street View, CelebA, Places2 and ImageNet.}
\label{tab0}
\end{table*}

\begin{figure*}[t!]
%\vspace{-0.05 in}
    \centering
    \setlength{\tabcolsep}{0.15em}
    {
        \begin{tabular}{ccccccc}
\includegraphics[width=2.2cm]{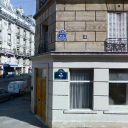}&
\includegraphics[width=2.2cm]{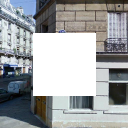}&
\includegraphics[width=2.2cm]{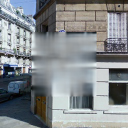}&
\includegraphics[width=2.2cm]{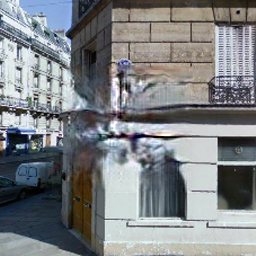}&
\includegraphics[width=2.2cm]{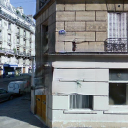}&
\includegraphics[width=2.2cm]{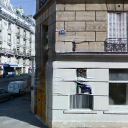}&
\includegraphics[width=2.2cm]{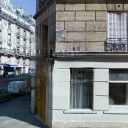}\\

\includegraphics[width=2.2cm]{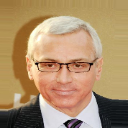}&
\includegraphics[width=2.2cm]{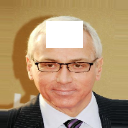}&
\includegraphics[width=2.2cm]{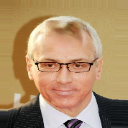}&
\includegraphics[width=2.2cm]{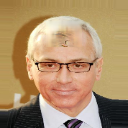}&
\includegraphics[width=2.2cm]{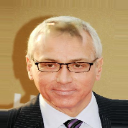}&
\includegraphics[width=2.2cm]{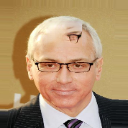}&
\includegraphics[width=2.2cm]{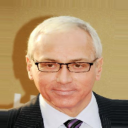}\\

\includegraphics[width=2.2cm]{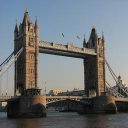}&
\includegraphics[width=2.2cm]{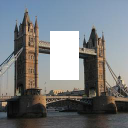}&
\includegraphics[width=2.2cm]{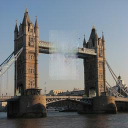}&
\includegraphics[width=2.2cm]{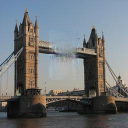}&
\includegraphics[width=2.2cm]{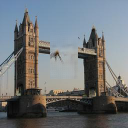}&
\includegraphics[width=2.2cm]{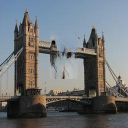}&
\includegraphics[width=2.2cm]{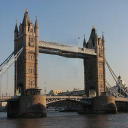}\\

\includegraphics[width=2.2cm]{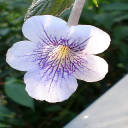}&
\includegraphics[width=2.2cm]{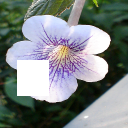}&
\includegraphics[width=2.2cm]{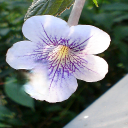}&
\includegraphics[width=2.2cm]{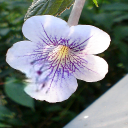}&
\includegraphics[width=2.2cm]{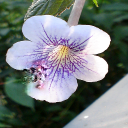}&
\includegraphics[width=2.2cm]{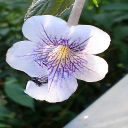}&
\includegraphics[width=2.2cm]{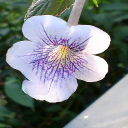}\\

\scriptsize(a) Original & \scriptsize(b) Corrupted &\scriptsize (c) CE &\scriptsize (d) GL  &\scriptsize (e) GntIpt &\scriptsize (f) GMCNN  &\scriptsize (g) Ours (GL+LID)\\

\end{tabular}  

}
    
\caption{Qualitative comparisons on Paris StreetView, CelebA, Places2 and ImageNet.}
\label{QualitativeCompare1}
% \vspace{-0.4cm}
\end{figure*}

% \textbf{Implementation Details}
We set $\lambda_{A}$ to $0.01$ as in \cite{yu2018generative,iizuka2017globally}.
Based on our analysis in Section \ref{ablative}, we empirically set $\lambda_{I} = 0.01$, $\lambda_{P} = 0.1$, $k_{I} = 8$ and $k_{P} = 5$ for our models.
All training images are resized and cropped to $256 \times 256$.
We choose \textit{conv4\_ 2} of VGG19 \cite{Simonyan2014Very} as the transformation $\phi$.
For all the tested models, we mask an image with a rectangular region that has a random location and a random size (ranging from $40 \times 40$ to $160 \times 160$).
The size of a training batch is 64.
The training takes 16 hours on Paris StreetView, 20 hours on CelebA and one day on both Places2 and ImageNet using an Nvidia GTX 1080Ti GPU.

\subsection{Quantitative Comparisons}
We compute the \textit{Peak Signal-to-Noise Ratio} (PSNR) and the \textit{Structural Similarity} (SSIM) between a restored image and its ground truth for quantitative comparison as in \cite{yu2018generative,Wang2018Image}.
As shown in Table \ref{tab0}, both of our proposed models CE+LID and GL+LID achieve better or comparable results to existing models across the four datasets, which verifies the effectiveness of submanifold alignment on different inpainting models.
Since GL+LID demonstrates more promising results than CE+LID, we only focus on the GL+LID model in the following experiments.

\subsection{Qualitative Comparisons}
We show sample images in this section to highlight the performance difference between our models and the four state-of-the-art models on the selected four datasets.
As shown in Figure \ref{QualitativeCompare1}, CE \cite{pathak2016context} can generate overall reasonable contents though the contents may be coarse and semantically inconsistent with surrounding contexts.
For example, the input image of the first row shows an building that has a part of the window missing. 
%However, the generated content by CE is inconsistent on this attribute.
CE can only generate coarse content with incomplete structures of the window and blurry details.
%CE also encounters segmentation misalignment for the images in the second (misplaced mouth part) and third (blurry facial edge) rows.
This is due to the limitation of the original GAN model that only provides overall high-level guidance (whether the inpaining output is generally similar to the ground truth image) and lacks specific regularization.
With the help of global and local discriminators, GL \cite{iizuka2017globally} can generate better details than CE but it still suffers from generating corrupted structures and blurry texture details.
Although GntIpt \cite{yu2018generative} can restore sharp details, it can lead to structural or textural inconsistency so that the restored contents may be contradictory with the surrounding contexts.
Since the first stage of GntIpt resembles the overall structure of GL, the coarse outputs from the first stage of GntIpt also have similar problems with GL.
Moreover, the second stage of GntIpt (finding the closest original patch from the surroundings with cosine similarity for the restored content) tends to easily attach simple and repeating patterns onto the restored content, thus even worsens the above mentioned problem.
Specifically in the first image, it fills the missing region with wall pattern ignoring the remaining window part at the bottom. 
GMCNN \cite{Wang2018Image} proposes an improved similarity measure to encourage diversity when finding the patch match, however it only insists on single patch diversity and thus causes unexpected artifacts and noises in the restored content.
In contrast, our model can generate more realistic results with better structural/textural consistency compared with the four state-of-the-art models.
%More visual inspections can be found in the  \href{https://www.github.com/ijcaiSubmissionDimIpt/ijcai_supplementary_material}{supplementary material}.

\begin{figure*}[t!]
%\vspace{-0.05 in}
    \centering
    \setlength{\tabcolsep}{0.2em}
    {
        \begin{tabular}{cccccc}
\includegraphics[width=2.7cm]{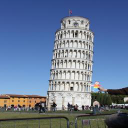}&
\includegraphics[width=2.7cm]{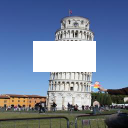}&
\includegraphics[width=2.7cm]{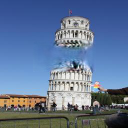}&
\includegraphics[width=2.7cm]{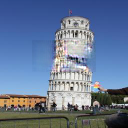}&
\includegraphics[width=2.7cm]{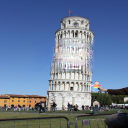}&
\includegraphics[width=2.7cm]{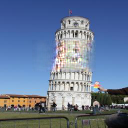}\\

% \small (a) Original & \small (b) Input & \small (c) $\lambda_{I} = 0$  & \small (d) $\lambda_{I} = 0.001$  & \small (e) $\lambda_{I} = 0.01$ & \small (f) $\lambda_{I} = 0.1$\\
%  & &\small \hspace{4mm} $\lambda_{P} = 0$ &\small \hspace{-1mm} $\lambda_{P} = 0$  &\small \hspace{0mm} $\lambda_{P} = 0$ &\small \hspace{1mm} $\lambda_{P} = 0$\\

\scriptsize (a) Original & \scriptsize (b) Corrupted & \scriptsize (c) $\lambda_{I},\lambda_{P} = 0,0$  & \scriptsize (d) $\lambda_{I},\lambda_{P} = 0.001,0$  & \scriptsize (e) $\lambda_{I},\lambda_{P} = 0.01,0$ & \scriptsize (f) $\lambda_{I},\lambda_{P} = 0.1,0$\\

\includegraphics[width=2.7cm]{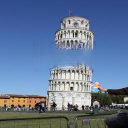}&
\includegraphics[width=2.7cm]{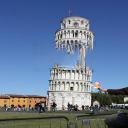}&
\includegraphics[width=2.7cm]{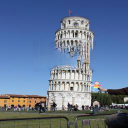}&
\includegraphics[width=2.7cm]{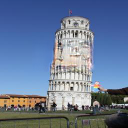}&
\includegraphics[width=2.7cm]{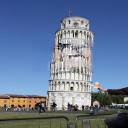}&
\includegraphics[width=2.7cm]{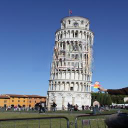}\\

% \small(f) $\lambda_{I} = 0$ &\small (g) $\lambda_{I} = 0$ &\small (h) $\lambda_{I} = 0$  &\small (i) $\lambda_{I} = 0.01$  &\small (j) $\lambda_{I} = 0.01$ &\small (k) $\lambda_{I} = 0.01$\\
% \small\hspace{8mm} $\lambda_{P} = 0.001$ &\small \hspace{8mm} $\lambda_{P} = 0.01$ &\small \hspace{7mm} $\lambda_{P} = 0.1$ &\small \hspace{5mm} $\lambda_{P} = 0.001$  &\small \hspace{4mm} $\lambda_{P} = 0.01$ &\small \hspace{3mm} $\lambda_{P} = 0.1$\\

\scriptsize (f) $\lambda_{I},\lambda_{P} = 0,0.001$ &\scriptsize (g) $\lambda_{I},\lambda_{P} = 0,0.01$ &\scriptsize (h) $\lambda_{I},\lambda_{P} = 0,0.1$  &\scriptsize (i) $\lambda_{I},\lambda_{P} = 0.01,0.001$  &\scriptsize (j) $\lambda_{I},\lambda_{P} = 0.01,0.01$ &\scriptsize (k) $\lambda_{I},\lambda_{P} = 0.01,0.1$\\
% \small\hspace{8mm} $\lambda_{P} = 0.001$ &\small \hspace{8mm} $\lambda_{P} = 0.01$ &\small \hspace{7mm} $\lambda_{P} = 0.1$ &\small \hspace{5mm} $\lambda_{P} = 0.001$  &\small \hspace{4mm} $\lambda_{P} = 0.01$ &\small \hspace{3mm} $\lambda_{P} = 0.1$\\

\end{tabular}  
    
}
    
    \caption{Qualitative comparisons of results with different combinations of $\lambda_{I}$ and $\lambda_{P}$.}
\label{QualitativeCompare2}
% \vspace{-0.5cm}
%\vspace{-0.15 in}
\end{figure*}

%\begin{table}[hbt]
%\vspace{-0.13 in}
%\caption{User preference for each model.}
%\label{vote}
%\renewcommand\arraystretch{1}
%\begin{center}
%\scalebox{0.9}
%{
%\begin{tabular}{lccccc}
%      			\hline
%      		Model & CE & GL & GntIpt & GMCNN  & Ours %(GL+LID)  \\ \hline
%      		Vote Counts & 5 & 5 &  43  & 168 & 739  \\ %\hline 
%\end{tabular}    
%}
%\end{center}
%\vspace{-0.27 in}
%\end{table}

%\subsection{User Study}
%We also conduct a user study for sanity check.
%We randomly choose 30 testing images, set random missing regions on them, then retrieve the restored images separately from each inpainting model.
%We invite 32 users to vote a single best result for each restored image, which gives us 960 votes in total.
%From the statistics in Table \ref{vote}, our model receives much more preference in terms of visual quality.

\begin{table}[htb]
%\vspace{-0.05 in}
\renewcommand\arraystretch{1}
\begin{center}
\scalebox{0.85}
{
\begin{tabular}{l|C|C|C|C}
      			\hline
      	 \multicolumn{5}{c}{$\lambda_{I} = 0$}  \\ \hline
     			 $\lambda_{P}$ & 0.001 & 0.01 & 0.1 & 1 \\ \hline 
		PSNR & 20.42  & 20.69  & \textbf{20.94}  & 20.46 \\	 
      		SSIM & 0.822  & 0.837  & \textbf{0.848}  & 0.820 \\ \hline \hline
      		
      		\multicolumn{5}{c}{$\lambda_{P} = 0$}  \\ \hline
     			 $\lambda_{I}$ & 0.001 & 0.01 & 0.1 & 1 \\ \hline 
		PSNR &  20.51 & \textbf{20.86}  & 20.80  & 20.54 \\	 
      		SSIM & 0.833  & \textbf{0.853}  & 0.844  & 0.830 \\ \hline \hline
      		
      		\multicolumn{5}{c}{$\lambda_{I} = 0.01$}  \\ \hline
     			 $\lambda_{P}$ & 0.001 & 0.01 & 0.1 & 1 \\ \hline 
		PSNR &  20.81 & 21.08  & \textbf{21.21}  & 20.84 \\	 
      		SSIM & 0.848  & 0.856  & \textbf{0.864}  & 0.843 \\ \hline \hline
      		
      		\multicolumn{5}{c}{$\lambda_{P} = 0.1$}  \\ \hline
     			 $\lambda_{I}$ & 0.001 & 0.01 & 0.1 & 1 \\ \hline 
		PSNR & 20.66  & \textbf{21.21}  & 21.02  & 20.75 \\	 
      		SSIM & 0.837  & \textbf{0.864}  & 0.858  & 0.839 \\ \hline
\end{tabular}    
}

\end{center}
% \vspace{-0.5cm}
%\vspace{-0.2 in}
\caption{Quantitative comparisons with different combinations of $\lambda_{I}$ and $\lambda_{P}$ on Places2.}
\label{tab1}
\end{table}

\begin{figure}[htb]
%\vspace{-0.1 in}
    \centering
    \setlength{\tabcolsep}{0.05em}
{
\begin{tabular}{cccc}
\includegraphics[width=2cm]{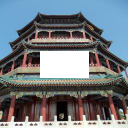}&
\includegraphics[width=2cm]{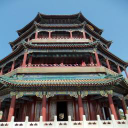}&
\includegraphics[width=2cm]{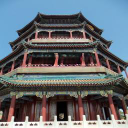}&
\includegraphics[width=2cm]{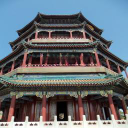}\\
\small(a) Corrupted &\small (b) $k_{P} = 1$ &\small (c) $k_{P} = 5$ &\small (d) $k_{P} = 9 $\\
\end{tabular}  
}
\caption{Qualitative comparisons on different $k_{P}$ values.}
\label{QualitativeCompare3}
%\vspace{-0.15 in}
\end{figure}

\subsection{Ablation Study}
\label{ablative}
\subsubsection{Parameter Analysis on iLID and pLID Regularizations}
We further conduct ablation experiments for our model on Places2.
Specifically, we investigate different combinations of iLID regularization and pLID regularization in the loss function by assigning $\lambda_{I}$ and $\lambda_{P}$ with different values in $[0.001, 1]$. 
An example of qualitative comparisons with different combinations of $\lambda_{I}$ and $\lambda_{P}$ is shown in Figure \ref{QualitativeCompare2}.
When $\lambda_{I}$ and $\lambda_{P}$ are both set to 0, the restored image encounters both structural and textural inconsistency problem with part of the building missing and filled with background (sky) pattern.
If we assign $\lambda_{I}$ with non-zero values and remain $\lambda_{P}$ as 0, the model becomes able to restore the overall structure of the building even if the texture details are noisy and blurry.
When we assign $\lambda_{P}$ with non-zero values and remain $\lambda_{I}$ as 0, the restored results have sharper and more diverse texture details on local patch level, but they fail to be consistent with the structure of original image on overall image level.
When we both assign $\lambda_{I}$ and $\lambda_{P}$ with non-zero values, we notice that the restored images are consistent on not only  overall structure but also texture details with the surrounding context.
From the visual results, we see that iLID regularization can increase structural consistency of restored contents, while pLID regularization focuses on patch-level alignment and thus benefits textural consistency.
%Specifically, we fix the tradeoff parameter $\lambda_{s}$ as 0.1 and train our model with variable values of tradeoff parameter $\lambda_{a}$.
%When $\lambda_{a}$ equals to zero which means only using segmentation regularization, the restored content is well-aligned with the surrounding segmentation structure, however almost no details can be observed.
%When $\lambda_{a}$ is small (\textit{e.g.}, $0.01$), the details of restored content is sharper but still blurry.
%When $\lambda_{a}$ becomes too large (\textit{e.g.}, $1$) which means the constrain is too excessive, extra artifacts and noises appear.
%Therefore, we empirically set $\lambda_{a} = 0.1$ in our experiments.
%The second group evaluates the effect of segmentation regularization as shown in Figure \ref{ablative2}.
%Accordingly we empirically set $\lambda_{s} = 0.1$ based on the comparisons.
Quantitative comparisons with 4 groups of parameter setting are given in Table \ref{tab1}.
We see that both iLID regularization and pLID regularization contribute to the model performance when using them alone.
Moreover, the collaboration of these two different regularizations can achieve complementary effects to improve inpainting quality.
We therefore empirically set $\lambda_{I} = 0.01$ and $\lambda_{P} = 0.1$ based on the ablation experiments.

% \textbf{The effect of different neighborhood sizes on iLID and pLID regularizations} \hspace{2mm}
\subsubsection{The Effect of Different Neighborhood Sizes}
We also investigate the effect of different neighborhood sizes on iLID ($k_{I}$) and pLID ($k_{P}$) regularizations.
For iLID regularization, different values of $k_{I}$ achieve similar results and empirically $k_{I} = 5 \sim 10$ strikes a better balance.
For pLID regularization, we find that assigning $k_{P}$ with different values can bring variable texture diversity for results.
Figure \ref{QualitativeCompare3} shows that $k_{P} = 5$ leads to more accurate texture details.
% (\textit{e.g.} more number of pillars).

\section{Conclusion}\label{secConclusion}
We studied the structural/textural inconsistency problem in image inpainting. 
To address this limitation, we propose to enforce the alignment between the local data submanifolds around restored images and those around the original images.
Our proposed inpainting model utilizes a combination of two LID-based regularizations: an image-level alignment regularization (with iLID) and a patch-level alignment regularization (with pLID) during the training process.
The experimental results confirm that our model achieves more accurate inpainting compared with the state-of-the-art models on multiple datasets.
%Our model can be improved further, for example, since our attribute and segmentation embedding networks are pretrained on auxiliary datasets, attribute vectors and segmentation maps might not be accurately predicted for input images, and this might result in an inaccurate supervision for inpainting.
We note that our enhancement can be applied to any GAN-based inpainting model, and further analysis in this direction can be interesting future work.

\section*{Acknowledgements}
The work is partially supported by the ARC grant DP170103174 and by the China Scholarship Council.

% \newpage
%% The file named.bst is a bibliography style file for BibTeX 0.99c
\bibliographystyle{named}
\bibliography{reference}

\end{document}